\documentclass[runningheads]{llncs}
\usepackage{graphicx}
\usepackage[numbers]{natbib}
\usepackage[utf8]{inputenc}
\usepackage{amsmath}
\usepackage{float}
\usepackage{amssymb}
\usepackage{xcolor}
\usepackage{bm}
\usepackage{hyperref}

\begin{document}
\title{Anomaly Detection for Skin Disease Images Using Variational Autoencoder 
\thanks{Code and data available at \url{https://github.com/QuindiTech/VAE_ISIC2018}}
}
%
%
\newcommand*\samethanks[1][\value{footnote}]{\footnotemark[#1]}

\author{
Yuchen Lu\inst{1,3}$^,$\thanks{Both authors contribute equally}, 
\and Peng Xu\inst{2,3}$^,$\samethanks
}
\authorrunning{Y. Lu et al.}
\institute{
Montreal Institute of Learning Algorithm (MILA), University of Montreal, Montreal QC, Canada \and
Computer Engineering, Polytechnique Montreal, Montreal QC, Canada \and
QuindiTech, Montreal, Canada\\
\url{http://www.quinditech.com},
}

\maketitle              

\begin{abstract}
In this paper, we demonstrate the potential of applying Variational Autoencoder (VAE) \cite{kingma2013auto} for anomaly detection in skin disease images. VAE is a class of deep generative models which is trained by maximizing the evidence lower bound of data distribution \cite{kingma2013auto}. When trained on only normal data, the resulting model is able to perform efficient inference and to determine if a test image is normal or not. We perform experiments on ISIC2018 Challenge Disease Classification dataset (Task 3)\cite{codella2018skin,DVN/DBW86T_2018} and compare different methods to use VAE to detect anomaly. The model is able to detect all diseases with 0.779 AUCROC. If we focus on specific diseases, the model is able to detect melanoma with 0.864 AUCROC and detect actinic keratosis with 0.872 AUCROC, even if it only sees the images of nevus. To the best of our knowledge, this is the first applied work of deep generative models for anomaly detection in dermatology. 
\end{abstract}

\section{Introduction}
Automatic skin disease detection would be valuable for both patients and doctors, and there has been success of applying deep supervised learning and CNN to the field of dermatology\cite{esteva2017dermatologist}. These models have large number of parameters and require large-scale labeled dataset for different kind of diseases. Nevertheless, human beings seem to be able to detect an abnormal skin lesion even if they are not trained, provided that they have enough experience with what a healthy mole looks like. Making our machine to have this behavior is interesting by itself, and it also provides practical advantages. By only observing normal skin image data, the algorithm is able to generalize to multiple diseases or even rare diseases, which saves time and money for data collection. Motivated by these aspects, we decide to focus on the problem of unsupervised anomaly detection for skin disease. Doing unsupervised learning over the space of images are challenging because of the curse of dimension, but recent development in deep generative models could address this issue. \\

There are two related models called Generative Adversarial Network (GAN) and Variational Autoencoder (VAE). 
Both VAE and GAN have been applied to anomaly detection \cite{kiran2018overview}. \cite{an2015variational} proposes using a direct ``reconstruction probability'' $E_{q(z|x)}\left[p(x|z)\right]$ for detection and shows VAE outperforms a PCA baseline on MNIST dataset. 
\cite{chen2018unsupervised} applies adversarial autoencoder to the unsupervised detection of lesions in brain MRI and improves the detection AUC for BRATS challenge dataset. 

Our major contribution is not proposing any fundamentally new methods, but to emphasize the potential usefulness of deep generative models in dermatology. We investigate VAE based methods instead of GAN for the following reason: 1) Even with recent tricks like gradient penalty, GAN training is still unstable and highly dynamic. As a contrast VAE training is more stable and therefore is more suitable as a proof of concept. 2) Most of GAN-based methods require training an additional network which maps from image space to the noise space in order to get the reconstruction \cite{kiran2018overview}, but it is unclear what the theoretical justification is of this additional network. On the contrary, VAE has a well defined mathematical framework and therefore is more interpretable.

\section{Methods}
We firstly give a brief introduction on the background of VAE and generative models. Then we propose different methods to use a trained VAE for anomaly detection.
\subsection{Variational Autoencoder}
VAE can be viewed as a directed probabilistic graphical model with the joint distribution defined as $p(\bm{x},\bm{z};\bm{\theta})=p(\bm{z})p(\bm{x}|\bm{z};\bm{\theta})$, where $\bm{x}\in\mathbb{R}^N$ is the data, $\bm{z}\in\mathbb{R}^M$ is the latent variable and $p(\bm{z})$ is the prior. We choose the prior to be $\mathcal{N}(0, I)$ in this work. When the true posterior $p(\bm{z}|\bm{x})$ is intractable, one can use a parametric distribution $q(\bm{z}|\bm{x};\bm{\phi})$ to approximate the posterior. Then in order to perform MLE, it is sufficient to maximize the evidence lower bound:
\begin{equation}
\log p(\bm{x}) \geq - KL(q(\bm{z}|\bm{x};\bm{\phi})||p(\bm{z})) + E_{\bm{z}\sim q(\bm{z}|\bm{x};\bm{\phi})}\left[\log p(\bm{x}|\bm{z};\bm{\theta}) \right]
\label{eqn:elbo}
\end{equation}
We choose $q(\bm{z}|\bm{x};\bm{\phi})=\mathcal{N}(\bm{\mu}_{enc}(\bm{x};\bm{\phi}),diag({\bm{\sigma}_{enc}^{2}(\bm{x};\bm{\phi}))})$ to be a Gaussian distribution with diagonal covariance, where $\bm{\mu}_{enc}, \bm{\sigma}_{enc}^{2}$ are the output of a neural network. Then by the reparameterization trick, the evidence lower bound becomes
\begin{equation}
\log p(\bm{x}) \geq E_{\bm{\epsilon}\sim \mathcal{N}(0, I)}\left[\log p(\bm{x}|\bm{z};\bm{\theta}) \right]- KL(q(\bm{z}|\bm{x};\bm{\phi})||p(\bm{z}))
\label{eqn:elbo_repara}
\end{equation}
where $\bm{z} = \bm{\mu}_{enc}(\bm{x};\bm{\phi}) + \bm{\epsilon} \bm{\sigma}_{enc}(\bm{x};\bm{\phi})$. Eqn. (\ref{eqn:elbo_repara}) is differentiable w.r.t. both \bm{$\theta$} and \bm{$\phi$}, and it can be trained from end to end. In this paper we choose $p(\bm{x}|\bm{z};\bm{\theta})\sim \mathcal{N}(\bm{\mu}_{dec}(\bm{z};\bm{\theta}), \sigma^{2}I)$ where $\sigma$ is pre-determined. Then maximizing Eqn. (\ref{eqn:elbo_repara}) is equivalent to minimizing
$$
E_{\bm{\epsilon}\sim \mathcal{N}(0, I)}\left[ \frac{(\bm{x} - \bm{\mu}_{dec}(\bm{z}))^T(\bm{x} - \bm{\mu}_{dec}(\bm{z}))}{2\sigma^2} \right] + KL(q(\bm{z}|\bm{x};\bm{\phi})||p(\bm{z}))
$$
One can observe that the function of $\sigma$ here is just adjusting the relative weight between reconstruction term and KL term, as a result, the final loss function to be minimized looks like
\begin{equation}
E_{\bm{\epsilon}\sim \mathcal{N}(0, I)}\left[ (\bm{x} - \bm{\mu}_{dec}(\bm{z}))^T(\bm{x} - \bm{\mu}_{dec}(\bm{z})) \right] + \beta KL(q(\bm{z}|\bm{x};\bm{\phi})||p(\bm{z}))
\label{eqn:objective}
\end{equation}
The resulting training objective can be viewed as a specific case of $\beta-$VAE, but our derivation is not from an optimization perspective like in \cite{higgins2016beta}.

\subsection{Anomaly Score}
The degree of anomaly can be characterized by the possibility of seeing $x$ appear under distribution $p(x)$. Therefore computing the anomaly score is essentially estimating $s(x) = -\log p(x)$.
Once we have a trained VAE, there are several ways to use it to generate an anomaly score $s(x)$ for the new image $x$. 
\subsubsection{VAE Based Score}
One choice is to use the negative of Eqn. (\ref{eqn:elbo}) as an anomaly score. That is
\begin{equation}
s_{vae}(x) = KL(q(z|x)||p(z)) - \frac{1}{L}\sum_{i=1}^L \log p(x|z_i) 
\label{eqn:s_vae}
\end{equation}
where $z_i\sim q(z | x)$. If $s_{vae}(x)$ is larger, then $x$ has higher loss and thus is more likely to be an outlier. Since we can decompose the loss into reconstruction term and KL term, we can just define the corresponding anomaly scores:
\begin{equation}
s_{vae}^{kl} = KL(q(z|x)||p(z))
\label{eqn:s_vaekl}
\end{equation}
\begin{equation}
s_{vae}^{reconst} = - \frac{1}{L}\sum_{i=1}^L \log p(x|z_i)
\label{eqn:s_vaer}
\end{equation}
The motivation of decomposition is to investigate how each term in VAE loss is useful for anomaly detection.

\subsubsection{Importance Weighted Autoencoder (IWAE) Based Score}
Importance Weighted Autoencoder \cite{burda2015importance} proposes a tighter lower bound on $\log p(x)$, which is
\begin{equation}
	\log p(x) \geq E_{z_1,...,z_K \sim q(z|x)}\left[ \log \frac{1}{K}\sum_{i=1}^K \frac{p(x|z_i)p(z_i)}{q(z_i|x)} \right]
\label{eqn:iwae_elbo}
\end{equation}
When $k=1$, we recover the ELBO used by VAE. When $k$ becomes larger, it's proved in \cite{burda2015importance} that the Eqn. (\ref{eqn:iwae_elbo}) would become a tighter bound than Eqn. (\ref{eqn:elbo}), resulting in a more accurate inference. Similarily we can use the negative of Eqn. (\ref{eqn:iwae_elbo}) to compute the anomaly score as 
\begin{equation}
s_{iwae}(x) = -\log \left(\frac{1}{L}\sum_{i=1}^L \frac{p(x|z_{i})p(z_{i})}{q(z_{i}|x)}\right)
\label{eqn:s_iwae}
\end{equation}
where $z_{i} \sim q(z|x)$. The corresponding KL score and reconstruction score are
\begin{equation}
s_{iwae}^{kl}(x) = -\log \left(\frac{1}{L}\sum_{i=1}^L \frac{p(z_{i})}{q(z_{i}|x)}\right)
\label{eqn:s_iwaekl}
\end{equation} 
\begin{equation}
s_{iwae}^{reconst}(x) = -\log \left(\frac{1}{L}\sum_{i=1}^L p(x|z_{i})\right)
\label{eqn:s_iwaer}
\end{equation} 
Although it is unclear whether a tighter lower bound estimate would help with outlier detection, we introduce these scores for the sake of comparison.
\section{Experiment}
\subsection{Model Architecture}
We use the architecture similar to DCGAN\cite{Radford2015umsupervised}. For the encoder, we avoid using linear layer to produce mean and log variance, but use two separate convolution layers. This architecture is fully convolutional and the number of convolution blocks are dependent on the input image size. In our implementation, the image size is 128, which makes the encoder consisted of 5 convolutional blocks and decoder consisted of 5 deconvolutional blocks respectively. ADAM is used as the optimizer with default setting. Hyperparameters are set as below.
\begin{itemize}
\item{$\beta$(weight for KL term): 0.01}
\item{learning rate: 1e-4}
\item{$L$(number of samples for calculating scores): 15}
\item{batch size: 32}
\item{training epochs: 40}
\item{latent dimension:300}
\end{itemize}

\subsection{Dataset and Proprocessing}
We use ISIC2018 Challenge dataset (task 3)\cite{codella2018skin,DVN/DBW86T_2018} which contains images from 7 diseases. A detailed dataset information can be found in Table \ref{table:isic_dataset}. For training the VAE, we use 6369 images as training set and 336 as validation set. For anomaly detection, we select 250 images from the validation set and 100 images from the rest of diseases. We normalize our data to have range from $-1$ to $1$ and resize each image to have size $128\times 128$. 
\begin{table}[]
\centering
\begin{tabular}{  l  c  c  c  c  c  c  c }
    \hline
    Disease & MEL & NV & BCC & AKIEC & BKL & DF & VASC \\ 
    \hline
    \# Images & 1113 & 6705 & 514 & 327 & 1099 & 115 & 142 \\ 
    \hline 
\\
\end{tabular}
\caption{ISIC2018 Challenge Task 3 Dataset}
\label{table:isic_dataset}
\end{table}

\begin{table}[]
\centering
\begin{tabular}{lccccccc}
                    & AKIEC          & BCC            & BKL            & DF             & MEL            & VASC           & All Disease    \\ \hline
$s_{vae}^{reconst}$      & \textbf{0.872} & \textbf{0.803} & 0.792          & \textbf{0.682} & 0.862          & \textbf{0.662} & \textbf{0.779} \\ 
$s_{iwae}^{reconst}$         & 0.871          & 0.802          & \textbf{0.793} & 0.678          & \textbf{0.864} & 0.657          & 0.777  \\ \hline

$s_{vae}^{kl}$           & 0.441          & 0.454          & 0.472          & 0.398          & 0.690          & 0.487          & 0.491          \\
$s_{iwae}^{kl}$     & 0.406          & 0.431          & 0.441          & 0.383          & 0.677          & 0.477          & 0.469          \\ \hline
$s_{vae}$          & 0.864          & 0.795          & 0.783          & 0.671          & 0.861          & 0.651          & 0.771          \\
$s_{iwae}$ & 0.864          & 0.795          & 0.784          & 0.670          & 0.861          & 0.648          & 0.771          \\ \hline 
\\
\end{tabular}
\caption{The AUC ROC Results of Disease Detection. For each column $x$, we show the AUC results of different anomaly scores when $x$ is the abnormal class. The last column is test against all diseases. The results is the average of 5 runs.}
\label{table:auc_results}
\end{table}

\begin{figure}
\centering
{\includegraphics[width=.45\linewidth]{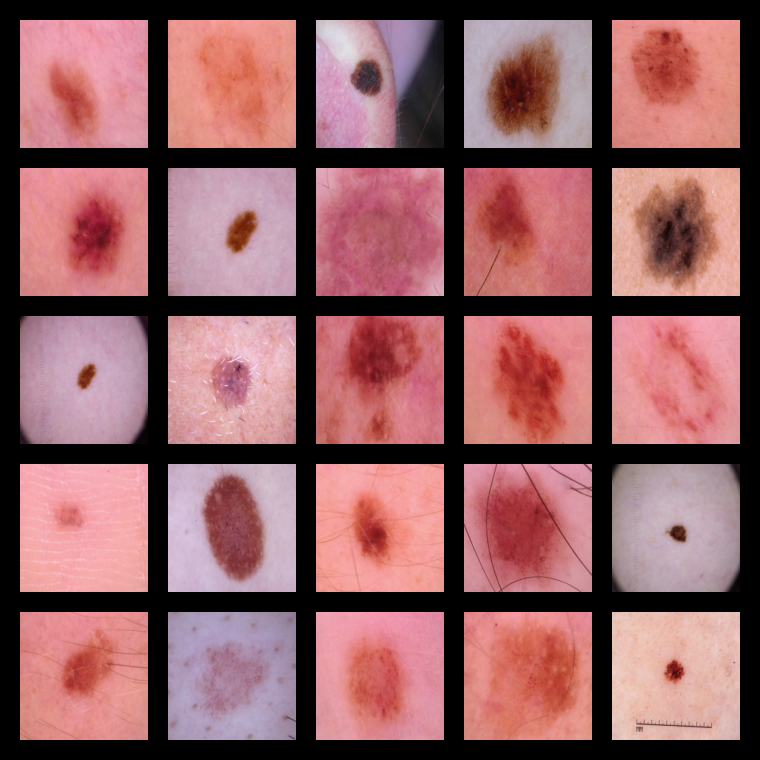}}
{\includegraphics[width=.45\linewidth]{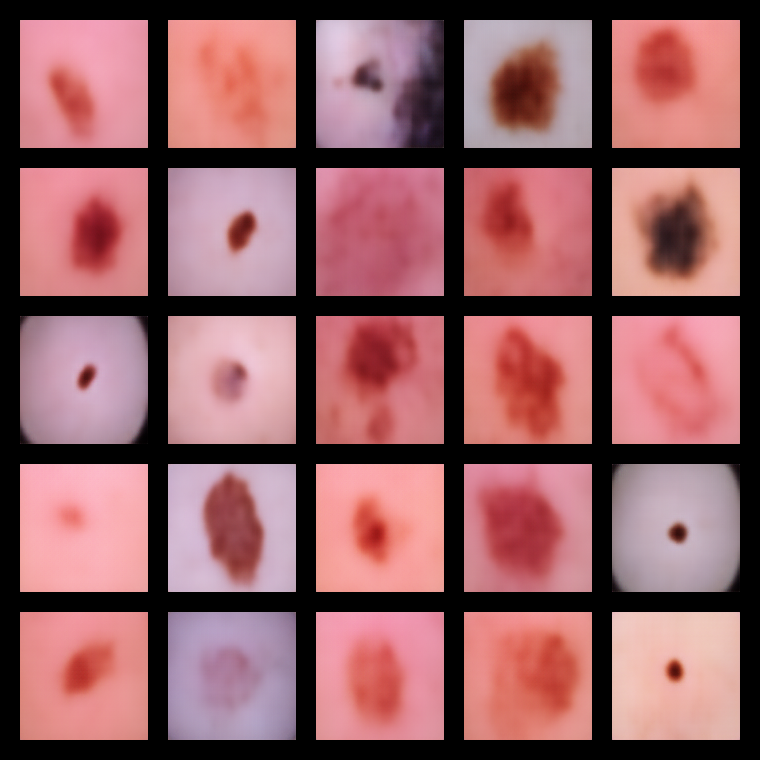}}
\caption[test]{\small{A non cherry-picked reconstruction result on validation set. \textit{left:} original images. \textit{right:} reconstruction images.}
}
\label{fig: reconstruction}
\end{figure}

The AUC result is summarized in Table \ref{table:auc_results}. Our best AUC result is obtained by reconstruction scores with an overall AUC score of 0.77.
In addition, the disease detection AUC for AKIEC and MEL reaches 0.87 and 0.86 respectively, even if the model has never seen a single image from these two diseases before. We notice that KL score is not very discriminative between normal and abnormal data. This is caused by using a small $\beta=0.01$ for the KL term, and model basically ignores the KL loss during training. We also try using a larger $\beta=1$ but it results in poorer AUC results. We also try using even smaller $\beta=0.001$, but it causes some numerical instability and the improvement is not significant. These results imply that the current prior is not expressive enough such that enforcing the approximated posterior $q(z|x)$ to be close to prior $p(z)$ hurts the model's expressiveness, which leads to worse AUC performance. We can also find that using IWAE variants scores does not make much difference from the VAE variants scores, which suggests that even if the bound is theoretically tighter\cite{burda2015importance}, its practical implication for anomaly detection might not be huge. A sample of reconstruction images is shown in Figure \ref{fig: reconstruction}.

\indent We try to compare our method with a traditional baseline like PCA or Kernel-PCA for anomaly detection, but our image size (3x128x128) is way too large for these methods to be implemented without resorting to feature engineering. This also demonstrate the advantage of using VAE to cope with the curse of dimensionality in anomaly detection.

\section{Future Work}
Based on our current experiment results, there are several future research direction worth pursuing.
\subsection{Improve VAE}  
As is shown above, our VAE faces the performance bottleneck because of the constraint to match posterior with a simple prior. One potential improvement would be adding a more expressive decoder like PixelVAE \cite{gulrajani2016pixelvae}. PixelVAE uses an expressive autoregressive structure for the decoder, which decomposes the lower level features from the higher level semantics. When the latent variable is only left to model the higher level feature, the simple Gaussian prior might be enough. From the reconstruction result, we can find the model is still outputting blurry images. This could be improved by using a more flexible posterior family 
or by doing hierachical variational inference \cite{sonderby2016ladder}.\\

\subsection{Improve Detection Methods}
In this work we haven't fully explored the method to use VAE for anomaly detection, but just use different outputs from VAE to compute the scores. One could fit a probability distribution (e.g. Gamma distribution) to the distribution of normal scores and use the standard statistical tests for anomaly detection. The latents of VAE can also be used for anomaly detection in several ways. One could train a one-class SVM using the latents as features. The latent space can also be used as a metric space so that the distance between two images can be defined by their inner product in the latent space. This enables one to develop a method similar to the metric learning based anomaly detection method \cite{du2014discriminative}.

\section{Conclusion}
In this paper we apply Variational Autoencoder (VAE) to the problem of anomaly detection in dermatology. VAE based anomaly detection method has a solid theoretic framework and is able to cope with high dimension data, like raw image pixels. Our objective is a specific case of $\beta-$VAE but from a different derivation. We experiment on ISIC 2018 Challenge Task 3 Dataset\cite{codella2018skin, DVN/DBW86T_2018}. By training only on normal data (nevus), the model is able to detect abnormal disease with 0.77 AUC. In particular, the model is able to detect AKIEC and MEL with 0.87 and 0.86 AUC respectively. This is to our knowledge the first work of applying Variational Autoencoder to dermatology, and we argue that although there have been successful applications of supervised learning and CNN based methods in dermatology, applying deep unsupervised learning in dermatology is a fruitful yet not fully explored research direction.
%
%
%
\bibliographystyle{splncs04}
\bibliography{reference.bib}


\end{document}